\documentclass[letterpaper, 10 pt, conference]{ieeeconf}  %

\IEEEoverridecommandlockouts                              %

\usepackage{amsmath} %
\usepackage{graphicx}

\usepackage{balance}

\usepackage[caption=false,font=footnotesize]{subfig} %

\usepackage{flafter} %

\usepackage{xcolor}
\usepackage{xspace}
\usepackage{cite}

\usepackage{algorithm}
\usepackage{algpseudocode}

\usepackage{caption}

\usepackage{booktabs}

\usepackage{float}

\usepackage[hidelinks,breaklinks]{hyperref}
\usepackage[capitalize,nameinlink]{cleveref}

\crefname{equation}{Eq.}{Eqs.}
\Crefname{equation}{Eq.}{Eqs.}

\newcommand{\newtask}[2]{%
  \expandafter\DeclareRobustCommand\csname #1\endcsname{%
    \texttt{#2}}%
}
\newtask{Square}{Square}
\newtask{Can}{Can}
\newtask{BoxCleanup}{BoxCleanup}
\newtask{CanSort}{CanSort}
\newtask{Coffee}{Coffee}
\newtask{WoollyBall}{WoollyBallPnP}
\newtask{Handover}{PackageHandover}

\newcommand{\newmethod}[3]{%
  \expandafter\DeclareRobustCommand\csname #1\endcsname{%
    \textcolor[HTML]{#2}{#3}\xspace}%
}

\newmethod{methodname}{000000}{ResFiT} %

\newcommand{\ish}{\ensuremath{\mathord{\sim}}\nobreak}

\let\OLDthebibliography\thebibliography
\renewcommand\thebibliography[1]{
  \OLDthebibliography{#1}
  \scriptsize
}

\usepackage[dvipsnames]{xcolor}
\newif\ifcomments
\commentstrue        %

\title{\LARGE \bf
Residual Off-Policy RL for Finetuning Behavior Cloning Policies
}

\author{%
\href{https://ankile.com}{Lars Ankile}\textsuperscript{*}$^{1,2}$,
\href{https://zhenyujiang.me/}{Zhenyu Jiang}$^{1}$,
\href{http://rockyduan.com/}{Rocky Duan}$^{1}$,
\href{https://www.gshi.me/}{Guanya Shi}\textsuperscript{\dag}$^{1,3}$,
\href{https://people.eecs.berkeley.edu/~pabbeel/}{Pieter Abbeel}\textsuperscript{\dag}$^{1,4}$,
and \href{https://scholar.google.com/citations?user=DkUUhXEAAAAJ\&hl=en}{Anusha Nagabandi}$^{1}$\\[0.4em]
\small $^{1}$Amazon FAR (Frontier AI \& Robotics) \quad $^{2}$Stanford University \quad $^{3}$Carnegie Mellon University \quad $^{4}$UC Berkeley
}

\begin{document}

\let\oldtwocolumn\twocolumn
\renewcommand\twocolumn[1][]{%
    \oldtwocolumn[{#1}{
    \vspace{-20pt}
    \begin{flushleft}
           \centering
    \includegraphics[clip,trim=0cm 0cm 0cm 0cm,width=0.99\textwidth]{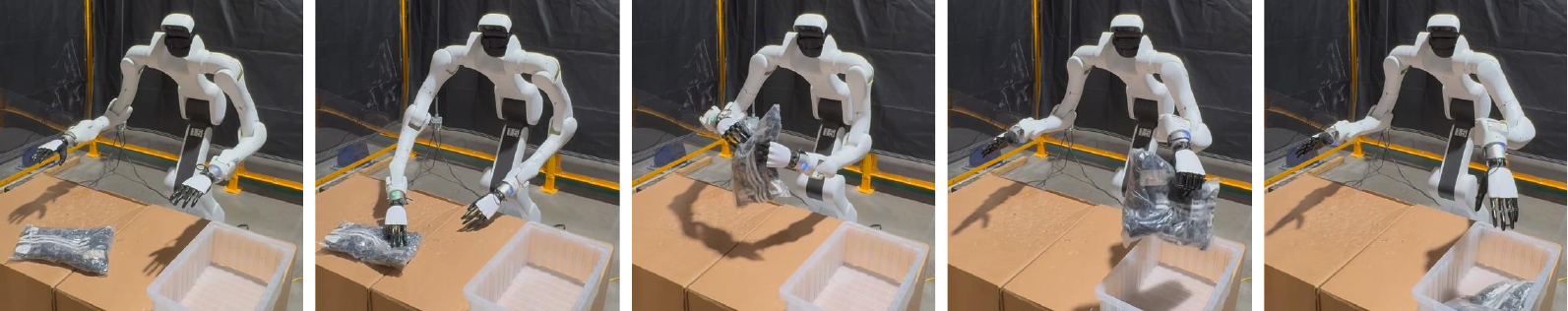}
    \captionof{figure}{%
    Our residual RL method (\methodname{}), performing real-world RL directly on our 29-degree-of-freedom (DoF) wheeled humanoid platform with two 5-fingered hands, is shown here performing the task of bimanual package handover.}\label{fig:teaser}
    \end{flushleft}
    }]
}

\begingroup
\renewcommand\thefootnote{\fnsymbol{footnote}}
\maketitle
\footnotetext[1]{Work done while an intern at Amazon FAR}
\footnotetext[2]{Work done while at Amazon FAR}
\endgroup

\thispagestyle{empty}
\pagestyle{empty}

\begin{abstract}
Recent advances in behavior cloning (BC) have enabled impressive visuomotor control policies. However, these approaches are limited by the quality of human demonstrations, the manual effort required for data collection, and the diminishing returns from offline data.
In comparison, reinforcement learning (RL) trains an agent through autonomous interaction with the environment and has shown remarkable success in various domains. Still, training RL policies directly on real-world robots remains challenging due to sample inefficiency, safety concerns, and the difficulty of learning from sparse rewards for long-horizon tasks, especially for high-degree-of-freedom (DoF) systems.

We present a recipe that combines the benefits of BC and RL through a residual learning framework. Our approach leverages BC policies as black-box bases and learns lightweight per-step residual corrections via sample-efficient off-policy RL.
We demonstrate that our method requires only sparse binary reward signals and can effectively improve manipulation policies on high-degree-of-freedom (DoF) systems in both simulation and the real world. In particular, we demonstrate, to the best of our knowledge, the first successful real-world RL training on a humanoid robot with dexterous hands.
Our results demonstrate state-of-the-art performance in various vision-based tasks, pointing towards a practical pathway for deploying RL in the real world.

\textcolor{cyan}{\href{https://residual-offpolicy-rl.github.io/}{Project website: residual-offpolicy-rl.github.io}}.
\end{abstract}

\section{INTRODUCTION}

Enabling robots to learn and improve directly in their deployment environments remains a fundamental challenge in robotics. Recently, significant progress has been made in training visuomotor control policies in the real world with behavior cloning (BC) from human demonstrations~\cite{brohan2022rt, bousmalis2023robocat, brohan_rt-2_2023, chi2023diffusion, zhao2023learning, zhao2024aloha, black2410pi0, bjorck2025gr00t, barreiros2025careful}. However, this success requires significant infrastructure, as well as numerous hours of manual and cumbersome data collection.
Even if unlimited data could be collected for every task, not only is human teleoperator performance generally suboptimal, but there is also emerging evidence that policy performance saturates with increasing demonstrations~\cite{ross_efficient_2010, zhao2024aloha, yu2024learning, ankile2024imitation, jiang2025dexmimicgenautomateddatageneration}.

Reinforcement learning (RL) offers a complementary paradigm: agents learn autonomously through trial and error in the environment. Deep RL has shown great success in various domains~\cite{mnih2015human, silver2016mastering, farooq2024survey, mirhoseini2021graph, guo2025deepseek, yu2025dapo, wang2024reinforcement, torne_reconciling_2024}, including in-hand manipulation~\cite{chen2022system, qi2023general} and locomotion~\cite{radosavovic2024real, gu2024advancing, margolis2024rapid, hwangbo2019learning}. However, strong RL performance generally requires large amounts of data from online interactions, so its application has been mainly in simulation~\cite{zhao2020sim, da2025survey} since real-world data are expensive and potentially unsafe to gather in large amounts.

A natural direction to improve BC policies is to leverage online RL~\cite{ankile2024imitation, mark_policy_2024, dong2025expo, yuan2024policy}, combining the strengths of each: BC policies provide a strong prior that can regularize exploration in the RL process, while online RL enhances policy performance by learning from interactions with the environment. However, modern BC architectures are typically deep models with tens of millions to billions of parameters that utilize action chunking or diffusion-based approaches, which can make it challenging to apply RL methods directly to optimize the policy. A simple yet powerful recipe that avoids several of the above issues is residual RL~\cite{silver2018residual, johannink2019residual, alakuijala2021residual, haldar2023teach_fish_residual, ankile2024imitation, yuan2024policy, dong2025expo}, where RL is applied not to learn a full policy, but only to learn corrective terms on top of a fixed base controller. Previous work has demonstrated that residual RL can indeed enhance the reliability of a pre-trained policy. Still, it has so far been limited to learning in simulation~\cite{ankile2024imitation, yuan2024policy, dong2025expo} or demonstrating results in simple or constrained settings~\cite{silver2018residual, johannink2019residual, alakuijala2021residual, haldar2023teach_fish_residual}; Applications to high-DoF systems learning directly in the real world are still lacking.

\begin{figure*}[t]
    \centering
    \includegraphics[width=\linewidth]{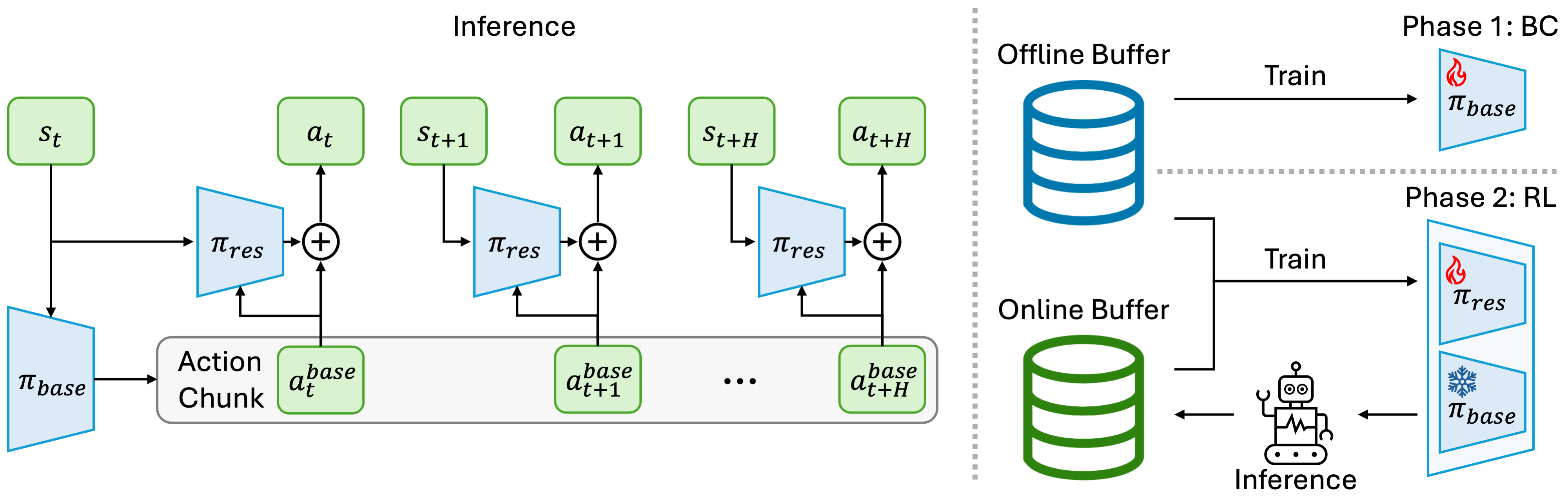}
    \caption{%
    Off-policy residual fine-tuning (\methodname{}): A two-phase approach using online RL to improve BC policies. First, we train a base policy using BC on a dataset of demonstrations and then freeze the base. Then, we learn a residual policy with RL to correct the base actions. The RL phase utilizes our off-policy recipe, leveraging both demonstrations and interactions, and enables more stable and safe exploration in the real world, as we can directly control the magnitude of the residuals. This residual approach is agnostic to the base policy parameterization and can be applied to large action-chunked BC policies to provide closed-loop corrections.
     }
    \label{fig:pipeline}
    \vspace{-15pt}
\end{figure*}

In this work, we present an off-policy residual fine-tuning (\methodname{}) approach that utilizes online RL to enhance BC policies. By treating the base policy as a black box and learning a per-step residual correction that is independent of chunk size and policy parameterization, we sidestep the challenges of directly optimizing huge base policies. By carefully designing our off-policy recipe, we make the RL process sample efficient enough to scale to high-DoF bimanual systems, require only sparse binary reward signals, and be safe enough to deploy in the real world. We demonstrate robust performance on sparse-reward, long-horizon, vision-based tasks, showing that our approach achieves state-of-the-art performance for a range of tasks in simulation. We also investigate each design decision in our recipe. To the best of our knowledge, we provide the first demonstration of RL on a humanoid robot with five-fingered hands, trained entirely in the real world.

\section{RELATED WORK}

\textbf{Behavior Cloning.}
BC has shown success in various settings, from autoregressive next-token prediction~\cite{brohan2022rt, radosavovic2024real, cui2022play, kim2024openvla} and diffusion policies~\cite{chi2023diffusion, team2024octo, zhao2024aloha, peebles2023scalable, black2410pi0}, to large transformers that directly output robot actions~\cite{zhao2023learning, kim2025fine, wang2024scaling}. Recent advances~\cite{intelligence2025pi05, bjorck2025gr00t, barreiros2025careful} leverage large transformers, diffusion and flow matching heads, and action chunking to handle long-horizon tasks and achieve impressive performance. Despite these advances, BC has fundamental limitations. The manual effort and infrastructure needed to scale up data collection to these levels are incredibly high, and recent work has shown diminishing returns and performance plateaus with increasing data~\cite{zhao2024aloha, jiang2025dexmimicgenautomateddatageneration, ankile2024imitation}.

\textbf{Fine-tuning BC Policies with RL.}
RL fine-tuning has been explored as a method to enhance BC policies by allowing the agent to interact directly with the environment. However, directly fine-tuning modern BC architectures with RL presents significant challenges. State-of-the-art BC policies typically use action-chunking or diffusion-based approaches~\cite{zhao2024aloha, chi2023diffusion} with large neural networks, which can lead to unstable learning when combined with conventional RL methods. Several works address these challenges~\cite{chen2025fdpp,hansen-estruch_idql_2023}. IBRL~\cite{hu2023imitation} trains an imitation policy and then uses it to propose actions for exploration and to bootstrap target values.
PA-RL~\cite{mark_policy_2024} sidesteps applying RL to complex model architectures by learning a Q-function to optimize actions instead of the policy.
Other approaches~\cite{ren2024diffusion, mcallister2025flow, lu2023contrastive} specifically adapt diffusion- or flow-based policies, such as DSRL~\cite{wagenmaker2025steering}, which runs RL over the latent-noise space of a frozen base diffusion policy and trains a policy to output that noise. Existing methods often couple the algorithm design with the BC policy structure, limiting flexibility across policy classes.

\textbf{Residual RL and Sample Efficiency.}
Residual RL~\cite{silver2018residual, johannink2019residual, alakuijala2021residual, haldar2023teach_fish_residual} provides an alternative approach of learning corrections on top of a fixed base policy, rather than optimizing the entire policy end-to-end.
Recent work, such as ResiP~\cite{ankile2024imitation}, combines a modern action-chunked BC policy with a single-step RL residual policy; however, its on-policy RL algorithm has too high sample complexity to be deployed in the real world.

Toward the goal of improved sample efficiency, prior work shows promising results through learning from demonstrations \cite{rajeswaran_learning_2018} and off-policy RL \cite{haarnoja_soft_2018, fujimoto2018addressing}; drawing from a host of these related works \cite{hester2018deep, ashvin2020accelerating, ball2023efficient}, we show in this work that demos can serve multiple purposes: (1) to pre-train a BC policy, (2) to warm up a critic, and (3) to remain in a buffer and be leveraged throughout the online RL phase.

Closest to our work, Policy Decorator~\cite{yuan2024policy} and EXPO~\cite{dong2025expo} demonstrate that off-policy RL can be used to train residual policies more efficiently. However, Policy Decorator learns residuals over entire action chunks rather than per-step corrections, while EXPO uses base policies without chunking. In both cases, the experiments are limited to simulation-only and single-arm tasks, and only state-based tasks for EXPO. In contrast, we demonstrate RL on tasks with higher degrees of freedom and longer horizons, and we show visuomotor policies on a bimanual humanoid robot in the real world.

\textbf{Real-World RL.}
In the category of real-world RL, QT-Opt\cite{kalashnikov2018scalable} utilized data from a fleet of 7 robots operating over the course of several weeks to train a robotic arm with a parallel jaw gripper to grasp diverse objects. More recently, SERL \cite{luo_serl_2024} demonstrated insertion and pick-and-place policies by combining a few demonstrations with less than an hour of online interaction. A follow-up work\cite{luo_precise_2024} demonstrated more complex behaviors and bimanual control by using a human-in-the-loop approach to overcome the exploration problem. Multiple works have demonstrated RL in the real world for single-arm robots with grippers \cite{seo2024continuous, hu2023imitation, wagenmaker2025steering, yang_robot_2023}, including \cite{dong_what_2025}, who showed that using Q-functions to guide batch online RL is more performant than imitation-based methods. On higher-DOF robots for locomotion \cite{radosavovic2024real, gu2024advancing} and manipulation \cite{jiang2025dexmimicgenautomateddatageneration, chen2025clutterdexgrasp}, people often learn in simulation and rely on sim-to-real transfer to run in the real world. For manipulation, CASHER~\cite{torne2024robot} trains policies in crowdsourced digital twins, achieving zero-shot transfer with sublinear scaling of human effort. However, it is limited to simple manipulation due to challenges in contact simulation.

\section{METHOD}

Our method of off-policy RL for residual fine-tuning (\methodname{}) of BC policies is illustrated in \autoref{fig:pipeline}. It takes as a starting point a base policy, typically trained with BC on an offline dataset of demonstrations using any algorithm or architecture. In our experiments, we chose a base policy with action chunking. Then, we perform online RL using our sample-efficient off-policy recipe to learn residual corrections on top of the base policy. In the following sections, we outline the details of the method, including the design choices and implementation details that are crucial to its success.

\subsection{Base Policy: Behavior Cloning with Action Chunking}

Consider an agent that receives observation $o_t$ and performs action $a_t$ at each timestep $t$. We first collect a dataset $\mathcal D_\text{demos}$ of demonstrations, e.g., through human teleoperation in the real world, consisting of successful trajectories $\tau = (o_0, a_0, o_1, a_1, \dots)$. Given this dataset, we train a base policy $\pi_\psi(a_{t:t+k}|o_t)$ using behavior cloning to predict a sequence of $k$ future actions at each timestep. We train the policy to maximize the log-likelihood of the action chunks from the demonstrations, i.e., $\min_\psi - \sum_{o_t, a_{t:t+k} \in \mathcal D_\text{demos}} \log \pi_\psi(a_{t:t+k}|o_t)$. 

Predicting a sequence of actions at each timestep, rather than a single action, is known to improve performance \cite{lai2022action, zhao2023learning, chi2023diffusion, ankile2024juicer} and alleviate compounding errors in imitation learning by effectively reducing the task horizon of the problem.

Note that although our policies receive only proprioception and image observations $o_t$ as input, and do not have access to the actual underlying system state $s_t$, the remainder of this section will use the state notation for simplicity.

\subsection{fine-tuning with Off-Policy Residual RL}

The above behavior cloning process produces a policy $\pi_\text{base}$ that has some level of task success. We freeze the base policy and train a new policy $\pi_\text{res}$ with RL to learn how to correct mistakes made by $\pi_\text{base}$ and improve policy performance in a way that is (1) agnostic to how $\pi_\text{base}$ is parameterized and trained and (2) stable since we can control the magnitude of this residual to stay relatively close to the base policy, especially during earlier exploration phases.

Consider a Markov decision process (MDP) \cite{bellman1957markovian} with states $s_t \in S$, actions $a_t \in A$, rewards $r_t = R(s_t,a_t)$, discount factor $\gamma$, and horizon $H$. RL aims to learn the parameters $\theta$ of some policy $\pi_\theta(s_t)$ such that the expected sum of rewards $R(\tau) = \sum_{t=0}^{H}{\gamma^t r_t}$ for a trajectory $\tau$ is maximized under the trajectory distribution induced by the policy. In this work, we will learn both a critic $Q_\phi$ as well as a policy $\pi_\theta$, as shown in DDPG \cite{lillicrap2015continuous}.

Whereas standard off-policy RL methods learn $Q_\phi(s_t,a_t)$ and $\pi_\theta(s_t)$, we reparameterize this for our residual setting as $Q_\phi(s_t, a_t^\text{base}+\pi_\theta(s_t, a_t^\text{base}))$ and $\pi_\theta(s_t, a_{t}^\text{base})$. In other words, the residual policy receives the current observation as well as the base action; the full action is the sum $a_t = a_t^\text{base} + a_t^\text{res}$, and the critic predicts the values of the full actions.

\begin{algorithm}
\caption{\methodname{}: Residual fine-tuning w/ Off-Policy RL}
\label{alg:main}
\begin{algorithmic}[1]
\Statex \textbf{Input:} pre-trained base $\pi_\text{b}$, dataset of demos $\mathcal D_\text{offline}$
\Statex \textbf{Initialize:} residual policy $\pi_\theta$, Q ensemble $Q_{\phi_1}, \dots Q_{\phi_N}$, vision encoder $f_\omega$, empty buffer $\mathcal D_\text{online}$
\Statex \textbf{Initialize:} set target networks $\theta' \leftarrow \theta$, $\phi'_i \leftarrow \phi_i$

\For {$\text{step} = 1,2,\dots, \texttt{buffer\_warmup\_steps}$}
    \State Sample noise $\epsilon_t \sim \mathcal{U}(-\text{noise\_scale},\text{noise\_scale})$
    \State Step env with $a_t = \epsilon_t + a_{t}^\text{b}$ where $a_{t}^\text{b} \sim \pi_\text{b}(s_t)$
    \State Observe next state $s_{t+1}$, reward $r_t$, done flag $d_{t}$
    \State Add transition $(s_t, a_{t}^\text{b}, a_t, s_{t+1}, a_{t+1}^b, r_t, d_t)$ to $\mathcal D_\text{online}$
\EndFor

\Repeat
    \State $a_{t}^\text{b} \sim \pi_\text{b}(s_t)$ and $a_{t}^\text{r} \sim \pi_\theta(s_t, a_{t}^\text{b})$
    \State Step env with $a_t = a_{t}^\text{b} + a_{t}^\text{r}$
    \State Add $(s_t, a_{t}^\text{b}, a_t, s_{t+1}, a_{t+1}^b, r_t, d_t)$ to $\mathcal D_\text{online}$
    \State Repeat UTD times:
    \State \hspace{1em} Sample batch $B$ evenly from $\mathcal D_\text{offline}$, $\mathcal D_\text{online}$
    \State \hspace{1em} Update critic using $B$:
    \State \hspace{2.5em}\makebox[0pt][r]{\color{black!40}\rule[-1ex]{0.4pt}{3ex}\hspace{0.2em}}$a_{{t+1}}^\text{r}\!\sim\!\pi_{\theta'}(s_{t+1}, a_{t+1}^\text{b})$
    \State \hspace{2.5em}\makebox[0pt][r]{\color{black!40}\rule[-1ex]{0.4pt}{3.5ex}\hspace{0.2em}}Compute $a_{t+1} = a_{t+1}^\text{b} + a_{t+1}^\text{r}$
    \State \hspace{2.5em}\makebox[0pt][r]{\color{black!40}\rule[-1ex]{0.4pt}{3.5ex}\hspace{0.2em}}$y = r_t + (1-d_t)*\gamma* \min_\text{subset(i)} Q_{\phi'_i}(s_{t+1}, a_{t+1})$
    \State \hspace{2.5em}\makebox[0pt][r]{\color{black!40}\rule[-1ex]{0.4pt}{3.5ex}\hspace{0.2em}}Update $\phi_i,\omega$ to minimize $\text{MSE} (Q_{\phi_i}(s_t, a_t),  y)$
    \State \hspace{2.5em}\makebox[0pt][r]{\color{black!40}\rule[-1ex]{0.4pt}{3.5ex}\hspace{0.2em}}Update critic targets $\phi'_i \leftarrow \rho \phi'_i + (1-\rho) \phi_i$
    \State Update actor using latest $B$:
    \State \hspace{1.2em}\makebox[0pt][r]{\color{black!40}\rule[-1ex]{0.4pt}{3ex}\hspace{0.2em}}Update $\theta$ to maximize: 
    \Statex \hspace{2.7em}\makebox[0pt][r]{\color{black!40}\rule[-1ex]{0.4pt}{3.5ex}\hspace{0.2em}} \hspace{1em} $\frac{1}{N} \sum_{i=1}^N Q_{\phi_i}(s_t, \pi_\theta(s_t, a_{t}^\text{b}) + a_{t}^\text{b})$
    \State \hspace{1.2em}\makebox[0pt][r]{\color{black!40}\rule[-1ex]{0.4pt}{3.5ex}\hspace{0.2em}} Update actor target $\theta' \leftarrow \rho \theta' + (1-\rho) \theta$
\Until{convergence}
\end{algorithmic}
\end{algorithm}

\begin{figure*}[t]
    \centering
    \begin{minipage}{0.79\linewidth}
        \includegraphics[width=\linewidth]{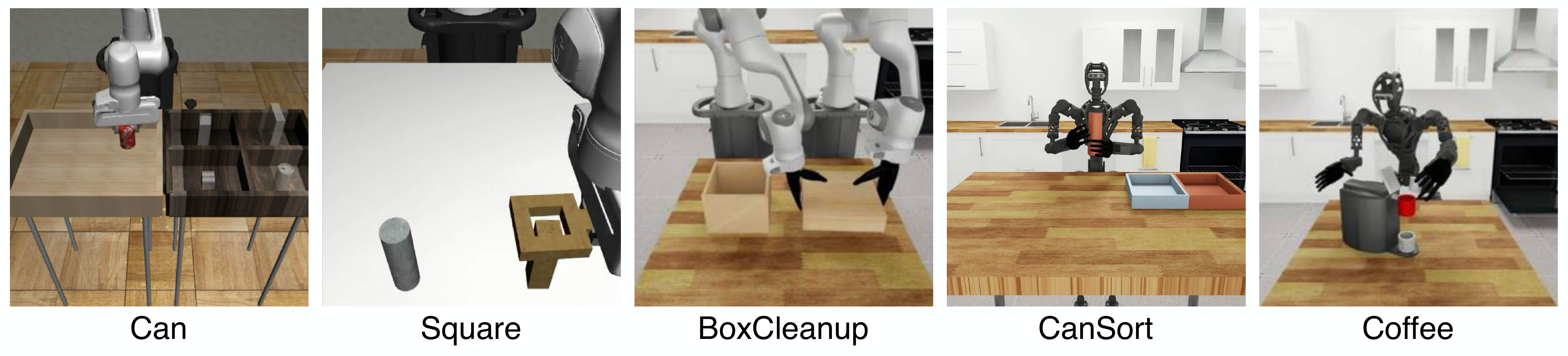}
        \caption{Our simulation tasks from Robomimic and DexMimicGen, spanning single-arm manipulation as well as bimanual coordination tasks.}
        \label{fig:task-overview}
    \end{minipage}%
    \hfill
    \begin{minipage}{0.18\linewidth}
        \includegraphics[width=\linewidth]{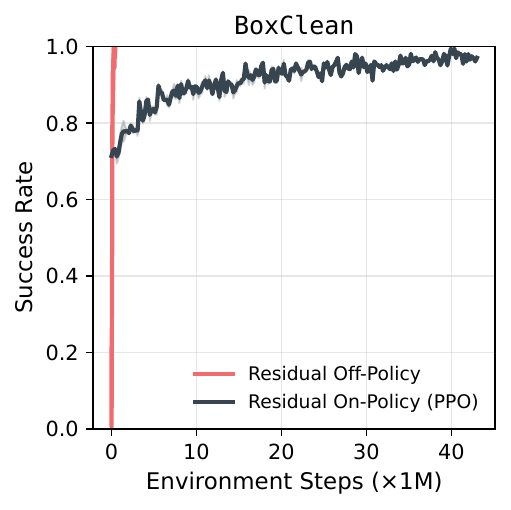}
        \caption{%
        PPO vs. off-policy RL in \methodname{}.
        }
        \label{fig:ppo}
    \end{minipage}
    \vspace{-10pt}
\end{figure*}

For the critic, recall the Bellman equation describing the optimal action-value function for a standard $Q^\star(s_t, a_t)$:
\begin{equation}
Q^\star(s_t,a_t) = \underset{s_{t+1} \sim P}{{\mathrm E}}\left[r_t + \gamma \max_{a'} Q^\star(s_{t+1}, a')\right].
\end{equation}
We train our value approximator $Q_\phi$ to approximate $Q^\star$ by using the Mean-Squared Bellman Error (MSBE) loss, which simply tells us how close our approximator is to satisfying the Bellman equation. Given a dataset {$\mathcal D$} of transitions $(s_t,a_t,r_t,s_{t+1},d_t)$, $s_{t+1}$ is the resulting state from taking action $a_t$ from state $s_t$, $r_t$ is the resulting reward, and $d_t$ indicates whether state $s_{t+1}$ is terminal. Using our policy in place of the $\max Q$, the loss for our residual setting becomes:
\begin{align}\label{eq:td-backup}
&L(\phi) = \underset{(s_t,a_t,r_t,s_{t+1},d_t) \sim \mathcal D} {\mathrm E} \left[ \Bigg( Q_{\phi}(s_t,a_t) - \right. \\ 
& \left. \left(r_t + \gamma (1 - d) Q_\phi(s_{t+1}, a_{t+1}^\text{base} + \pi_\theta(s_{t+1}, a_{t+1}^\text{base})) \right) \Bigg)^2 \right] 
\nonumber 
\end{align}
where $a_{t+1}^\text{base} = \pi_\text{base}(s_{t+1})$.
Moving on to the policy, recall that given an optimal action-value function $Q^\star(s,a)$, the optimal action $a^\star$ can be found by simply taking the max over actions of $Q^\star(s,a)$. 
So in our case of continual actions, and given that Q is differentiable, we can train our policy $\pi_\theta(s)$ by simply performing gradient ascent on the value function, with respect to the policy parameters:
\begin{equation}
L(\theta) = - \underset{\left(s_t, a_{t}^\text{base}\right) \sim {\mathcal D}}{{\mathrm E}}\left[ Q_{\phi}(s_t, a_{t}^\text{base} + \pi_{\theta}(s_t, a_{t}^\text{base})) \right].
\end{equation}

\subsection{Design Decisions}
\label{sec:tricks}

As described above, our approach uses off-policy data and interleaves using the Bellman equation to train the Q-function with using the Q-function to train the policy. In the following, we detail the key design choices that were required to achieve stable residual fine-tuning performance across different tasks and embodiments, present our complete method in Algorithm \ref{alg:main}, and later validate these decisions through ablations.

We use Update-to-Data ratio (UTD) $>1$, as prior works have shown that increasing the UTD, i.e., the number of model updates one takes per data point collected in the environment, can be an effective way of improving sample efficiency~\cite{ball2023efficient, chen2021redq}.
We also use $n$-step returns (in particular, our $n=3$), similar to CQN~\cite{seo2024continuous} and IBRL~\cite{hu2023imitation}, which is shown to help with long-horizon and sparse-reward tasks~\cite{park2025horizon}. Whereas Monte Carlo methods roll out the entire trajectory before doing updates, and standard Q-learning looks ahead just 1 step with value bootstrapping $r_t + \gamma Q(s_{t+1},a_{t+1})$, here we instead look ahead $n$ steps: $\sum_{i=0}^{n-1} \gamma^{i} r +  \gamma^n Q(s_{t+n},a_{t+n})$\cite{Sutton1998}. Note that we omit this detail in Algorithm \ref{alg:main} for simplicity. 

A common issue for off-policy RL is that when the Q-function gets queried with out-of-distribution actions, the values can often be overestimated due to the use of function approximation. Drawing from RLPD \cite{ball2023efficient}, we add layer normalization \cite{ba2016layernormalization} to the critic to mitigate catastrophic overestimation without explicitly constraining the policy, which turns out to be crucial to the performance of the policy.

We also incorporate several other established practices in our implementation.
Following TD3~\cite{fujimoto2018addressing}, we employ delayed actor updates (every 2-8 critic updates) to alleviate instability that stems from updates done with poor value estimates, we use Polyak averaging ($\tau=0.005$) to update target networks, and we perform target policy smoothing by adding noise to the action that forms the Q target, to help with brittle behavior caused by sharp peaks in learned Q function approximators.
We also use Randomized Ensembled Double Q-Learning~\cite{chen2021redq} for reducing overestimation bias, by computing TD-targets as $y = r + \gamma \min_{i \in S} Q_{\phi_i}(s', \pi_\theta(s'))$ from a random subset $S$ of $N$ Q functions, while policy updates utilize the full ensemble: $\nabla_\theta \frac{1}{N} \sum_{i=1}^N Q_{\phi_i}(s, \pi_\theta(s))$.
For visual inputs, we adopt a shallow ViT~\cite{dosovitskiy2020image} encoder with DrQ-style~\cite{kostrikov2021imageaugmentation} random shift augmentations to help with overfitting. 
Finally, we use our demonstration data not only to train our base policy but also during the online RL phase via symmetric sampling~\cite{ball2023efficient}, by sampling 50\% of each batch from this frozen offline data, and the remaining 50\% from our constantly growing online buffer.

\section{EXPERIMENTAL SETUP}

\begin{figure*}[t]
  \centering
    \includegraphics[width=1.0\linewidth]{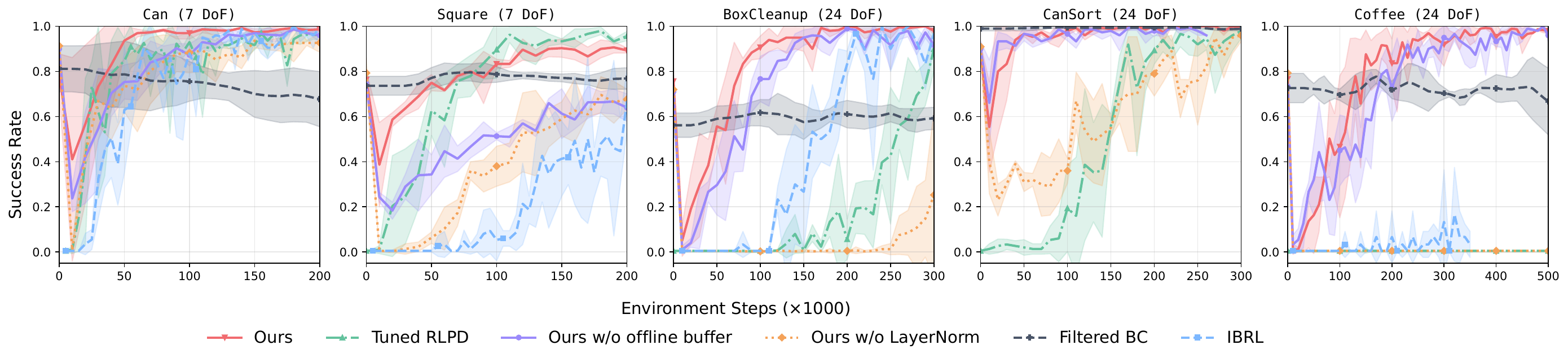}
  \caption{Success rates of different approaches on our simulation tasks, showing \methodname{} converging to high-performing policies for all tasks. Note that all approaches here start with the same number of demos.}
  \label{fig:learning-curves}
  \vspace{-15pt}
\end{figure*}

\subsection{Simulation Experiments}

We evaluate \methodname{}'s real-world tractability via simulation experiments using realistic constraints: a single simulation environment, image and robot joint state observations (no privileged object state information), and sparse binary rewards. We test our approach on tasks from Robomimic~\cite{mandlekar2021matters} and DexMimicGen~\cite{jiang2025dexmimicgenautomateddatageneration}, which vary in terms of robots, control modes, degrees of freedom, task horizons, and precision requirements.
Our experiments utilize the Robosuite environment~\cite{robosuite2020}, which is built on MuJoCo~\cite{todorov2012mujoco}, and we operate at a control frequency of 20 Hz.

\textbf{Tasks.} Our task selection (see Fig. \ref{fig:task-overview}) spans two categories: single-arm manipulation tasks and bimanual coordination tasks. For single-arm, we use \Can{} and \Square{} from Robomimic. These tasks use a Franka robot with a parallel-jaw gripper and have an action space dimension of 7, 6 for the delta end-effector pose, and 1 for the gripper action. To demonstrate the versatility of \methodname{}, we also include the bimanual \BoxCleanup{}, \CanSort{}, and \Coffee{} tasks from DexMimicGen, which require coordinated control of two arms with 6-DoF dexterous hands. \BoxCleanup{} uses dual Frankas to coordinate picking up a box lid and placing it precisely on top of the box, testing spatial coordination and alignment. \CanSort{} and \Coffee{} both use a GR1 humanoid robot with a fixed base; \CanSort{} entails picking up and passing a cylinder between its hands, and \Coffee{} requires dexterously grasping a coffee pod and placing it precisely into a coffee maker before closing the lid with the other hand. The bimanual tasks have action spaces with 24 dimensions: 6 for the absolute end-effector pose per arm, and 6 for the actuated joint positions per hand. Note that the end-effector control modes differ between these robots, dual Frankas use delta pose control (same as single-arm), while the GR1 uses absolute pose control. For all tasks, we use the released demonstration datasets with 300 demos from the Multi-Human dataset per single-arm task and 1000 demos created with DexMimicGen per bimanual task.

\textbf{Baselines and Ablations.} First, we examine how off-policy residual RL on top of an action-chunked base policy compares to directly performing off-policy RL to learn a single-action policy, starting from the same demos. For this, we chose the state-of-the-art algorithm RLPD~\cite{ball2023efficient}, an off-policy approach that includes both offline demos and online data in each training batch, and uses layer norm to help alleviate the overestimation of the value function. However, with default settings, it is unable to solve our high-DoF tasks when learning from images. Instead, we create an optimized version of RLPD by incorporating the same off-policy RL design decisions (see Sec.~\ref{sec:tricks}) that we use in our method. We refer to this off-policy RL implementation as ``Tuned RLPD.'' This means that the same number of demos are used, the same off-policy RL algorithm is used, and the same vision encoders and network architectures are used; the main difference is whether a base policy is being used. For another baseline of using demos but not performing residual learning, we compare to IBRL~\cite{hu2023imitation}, which uses a pre-trained BC policy during RL to propose actions and bootstrap target values. 
Finally, we include an online BC fine-tuning baseline called ``Filtered BC,'' which starts with the same base policy as \methodname{}, but does not use residual RL to perform the fine-tuning. It fine-tunes by iteratively performing rollouts and adding successes back into the dataset for continued BC training. This approach is similar in spirit to reward-weighted regression~\cite{peters2007reinforcement} (with 0/1 weights) and self-imitation~\cite{oh2018self, riedmiller_collect_2021}, which prior work has shown can improve performance over the demo-only case~\cite{bousmalis2023robocat, ankile2024juicer, lampe_mastering_2023, wang2024lessonslearningspinpens}.

Other design decisions that we ablate include whether we use the layer norm for the actor and critic MLPs, whether we incorporate the demos during the online RL phase, the choice of the UTD ratio, and the choice of $n$ for $n$ step returns used in the TD targets. Finally, we also compare to the more common version of residual RL\cite{ankile2024imitation}, which is to use online RL (PPO) to learn residuals.

\subsection{Real-World Experiments}

\begin{figure*}[t]
  \centering
  \includegraphics[width=0.495\textwidth]{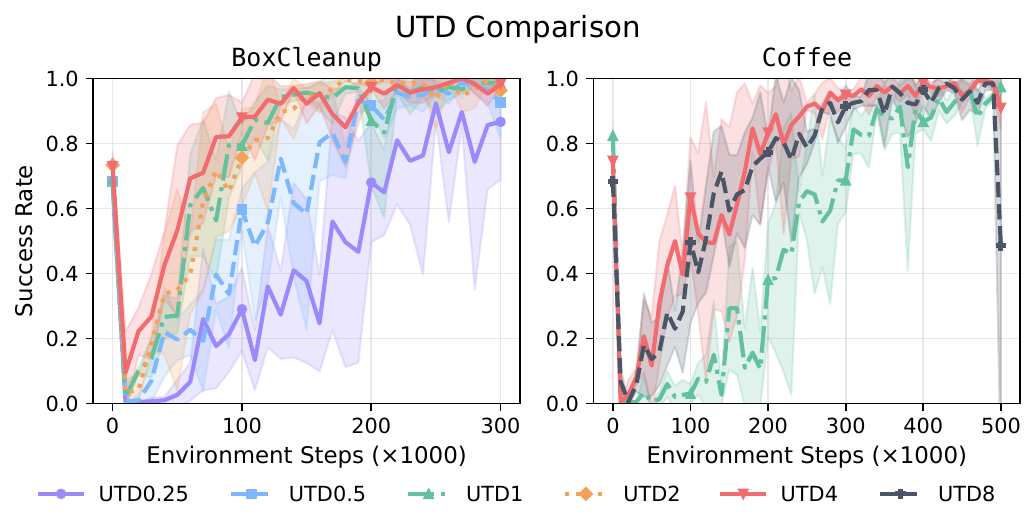}\hfill
  \includegraphics[width=0.495\textwidth]{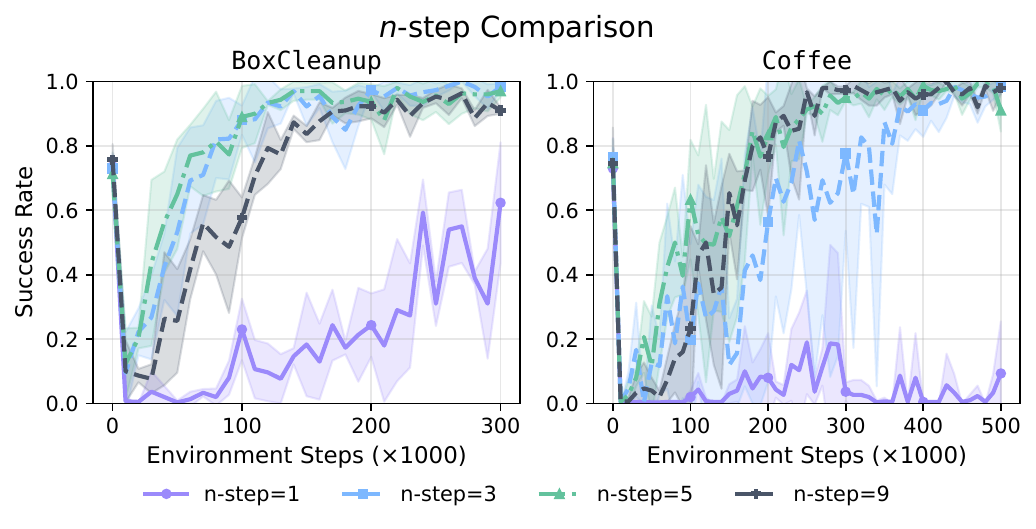}
  \caption{Impact of UTD ratio (left) and \texttt{n-step} (right) on the \BoxCleanup{} and \Coffee{} tasks.}
  \label{fig:ablations-2x2}
  \vspace{-10pt}
\end{figure*}

We demonstrate \methodname{} on a real-world bimanual dexterous manipulation platform - the wheeled Vega humanoid from Dexmate. Vega has two 7-DoF arms, two 6-DoF OyMotion dexterous hands, and 2 image streams coming from a Zed camera mounted on its 3-DoF head. We use absolute joint position control in the real world for each of these joints, making the action space dimension 29.

\textbf{Tasks.} We perform two tasks in the real world: \WoollyBall{} and \Handover{}. In \WoollyBall{}, the right hand picks up a ball from a random table location and puts it into a randomly placed tote. In \Handover{}, the right hand picks up a deformable package, hands it to the left hand, and places it into a tote on the left side of the workspace.

\textbf{Real-world Infrastructure.} Our real-world setup includes safety limits implemented using the force-torque sensor in the robot's wrists, and a collision checking implementation to prevent robot self-collisions. We built a system for performing RL in the real world, where the operator either lets each episode timeout or marks the result of each episode as success or failure, and then resets the scene. During RL training, we noticed that the bottleneck in wall-clock time was due to the model updates, as opposed to the data collection, due to the high UTD ratio. To enable higher data throughput, we split the actor and the learner into separate processes. In particular, after a completed trajectory that either succeeded, failed, or timed out, the learner process loads the latest replay buffer and performs model updates, while the data collector resets the scene and starts the next trajectory with the actor process. Note that the only reward for these rollouts was a single $1$ for success and $0$ otherwise.

\textbf{Evaluation protocol.} Real-world robot evaluations often suffer from confounding factors like environmental drift and lighting changes, especially when policies are evaluated sequentially rather than in matched conditions \cite{barreiros2025careful, kressgazit2024robotlearningempiricalscience}. To address these issues and fairly quantify performance differences between pre-trained and fine-tuned policies, we use blind A/B testing with matched initial conditions. For each evaluation round, we (1) sample a random scene configuration (object, tote), (2) randomly assign each policy to labels A and B, (3) execute both policies from identical initial states, and (4) reveal policy identities after completion. This approach attempts to mitigate evaluator bias, controls for sensitivity to initial conditions, and allows both policies to face similar environmental conditions within each round.

\section{RESULTS}

\subsection{Simulation Results}
 
We evaluate various aspects of \methodname{} on simulated benchmark tasks, consisting of both single-arm and bimanual manipulation tasks. On the simulated \BoxCleanup{} task, we first compare our off-policy residual RL recipe to an existing approach~\cite{ankile2024imitation} of performing residual RL in sim with the on-policy PPO \cite{schulman2017proximal} algorithm. As shown in \autoref{fig:ppo}, our approach converging at 200k steps versus 40M steps, we see a \ish$200\times$ boost in sample efficiency, demonstrating the need for off-policy approaches when considering performing the RL directly in the real world.

Examining the main simulation results in \autoref{fig:learning-curves}, we first observe that \methodname{} converges to near-perfect policies for all tasks. We find that for the simplest task \Can{}, the baseline off-policy methods as well as the ablated versions of our method all achieve high success rates in \ish150k steps, while \methodname{} converges faster, at \ish75k steps. 
In the \Square{} task, only \methodname{} and our optimized off-policy approach (``Tuned RLPD''), which does not use a base policy or residuals but does incorporate all of \methodname{}'s other off-policy RL design decisions, can achieve over 90\% success rate in 150k steps. 

For the harder tasks with more DoFs and longer horizons, like \BoxCleanup{}, \CanSort{}, and \Coffee{}, baselines and ablated versions either collapse to zero success rate or take longer to achieve high performance, while \methodname{} can still converge to near-perfect task performance efficiently. On \Coffee{}, which has the longest task horizon while still requiring the most precision, we notice that not having demos during the online RL of \methodname{} phase may not matter, but all of the approaches without action-chunking fail in this sparse reward and vision-based setting.

Across all tasks, the Filtered BC baseline remained stable but showed minimal improvement over the initial BC policy performance. We hypothesize that this is because the main failure mode is precision, which is hard to gain without explicit value maximization. In general, we find that Filtered BC is able to improve BC policies that start from a lower initial performance (e.g., \ish20\%), but quickly saturate, which is supported by previous work~\cite{ankile2024juicer}.

In \autoref{fig:ablations-2x2}, we study the effect of UTD and \texttt{n-step} returns for sparse reward tasks. Here, we see the importance of using an \texttt{n-step} larger than 1 for tasks where rewards are sparse. However, since a larger \texttt{ n-step} also increases bias, a too large value can affect performance. For UTD ratios, we find that for a task with a horizon length of 150-250 steps, such as \BoxCleanup{}, UTD values greater than 1 provide clear benefits, but gains plateau at moderate values. Specifically, learning is noticeably slower for a UTD of $\frac{1}{2}$, but increasing to very high UTDs of 8 or higher, as in some prior work, yields diminishing returns, with UTDs of 4 already capturing most of the benefit while still being stable.

\begin{figure*}[t]
  \centering
  \includegraphics[width=1.0\textwidth]{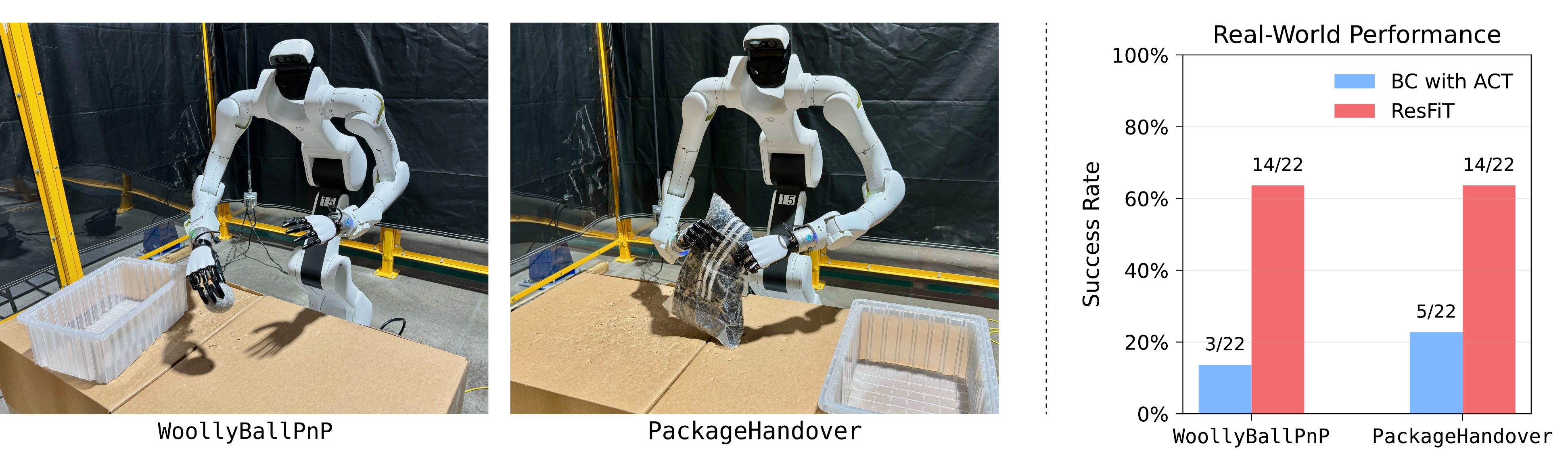}
  \caption{Real-world results of applying \methodname{}, where our residual RL approach shows a significant boost in performance over the base model, on a 29-DoF bimanual wheeled humanoid robot with two 5-fingered hands.}
  \label{fig:results_wooly_ball}
  \vspace{-10pt}
\end{figure*}

\subsection{Real-World RL Results}

In the real world, we apply our methods to two tasks, \WoollyBall{} and \Handover{}. We use ACT~\cite{zhao2024aloha} as the base policy for both tasks. 

For \WoollyBall{}, we trained our base ACT policy on a pick-and-place task with around 1,000 demonstrations across 4 different objects with random robot start locations, random object locations, and random placement tote locations. We then evaluated it on the object that it most struggled with: a small, gray, and woolly ball. The base policy achieved only 14\% success on this task, with the main failure mode being hovering and missed grasps with the dexterous hand. With 134 rollouts of autonomous RL execution ($\approx15$ minutes worth of robot execution data) specifically on this woolly ball, \methodname{} boosted the performance of the base model from 14\% to 64\%.

\Handover{} is a more difficult task since it requires two-arm coordination and involves a longer task horizon before experiencing any potential reward. For this task, our base policy was able to achieve a 23\% success rate with around 900 demonstrations, and after 343 RL episodes ($\approx76$ minutes worth of data) in the real world, \methodname{} was able to boost the task performance to 64\%. To the best of our knowledge, this is the first demonstration of real-world RL performed on a bimanual dexterous manipulation humanoid with two five-fingered hands, trained fully in the real world.

\section{DISCUSSION}

We demonstrate that decoupling the pre-training and fine-tuning stages can resolve a tension in the current robot learning paradigm: action chunking and larger policies improve BC performance but create intractably high-dimensional action spaces for RL (870 dims in our case). By learning only single-step corrections atop a frozen base policy, we maintain long-horizon reasoning while keeping optimization tractable.

An insight from our experiments is that the base policy serves a dual purpose beyond providing a good initialization. First, it acts as an implicit safety constraint: Policies trained without this regularization converge to faster but more aggressive behaviors that are unsuitable for real-world deployment. Second, it provides a powerful exploration prior that enables RL from sparse rewards even in high-DOF spaces. This suggests that hybrid BC-RL approaches are promising for high-DoF and long-horizon manipulation.

The approach's primary limitation is that learned behaviors may remain constrained around the base policy. Although our method clearly improves success rates in both simulation and real experiments, it cannot discover fundamentally different strategies or skills beyond what the base policy already encodes while remaining stable overall.

Looking ahead, figuring out the right way to relax the frozen base constraint without sacrificing stability could provide further improvement in performance and robustness. If we can distill the more precise, reliable, and fast behavior from the combined policy back into the base policy, that would provide more room for the residual model to improve further. This would be particularly powerful in the multitask setting, where one can distill task-specific residual improvements into an increasingly capable generalist.

Our real-world deployment validates that sample-efficient RL can work on bimanual platforms with 5-fingered hands. However, our experiments still required human supervision for resets and reward labeling. Without automatic reset mechanisms, success detection, and safety rails, even sample-efficient RL methods cannot enable autonomous skill improvement that scales independently of human oversight.

\section*{ACKNOWLEDGMENTS}

We thank Himanshu Gaurav Singh for technical discussions, helping with the filtered BC baseline, and feedback on the paper draft. Younggyo Seo provided valuable input for real-world implementation and offered constructive feedback on the paper. We also thank Carmelo Sferrazza and Ruihan Yang for their comments on the manuscript.
Benjamin Colby, Hassan Farooq, and Sumedh Sontakke helped with real-world experiments.
Finally, we appreciate the technical discussions during early project development from Quiang Li, Seohong Park, Perry Dong, Hengyuan Hu, and Suvir Mirchandani.

\newpage

\bibliographystyle{IEEEtran}
\bibliography{references}  %

\end{document}